# Joint Characterization of Multiscale Information in High Dimensional Data

Daniel Sousa[1]

Christopher Small
Lamont-Doherty Earth Observatory
Columbia University

Abstract

High dimensional data can contain multiple scales of variance. Analysis tools that preferentially operate at one scale can be ineffective at capturing all the information present in this cross-scale complexity. We propose a multiscale joint characterization approach designed to exploit synergies between global and local approaches to dimensionality reduction. We illustrate this approach using Principal Components Analysis (PCA) to characterize global variance structure and t-stochastic neighbor embedding (t-sne) to characterize local variance structure. Using both synthetic images and real-world imaging spectroscopy data, we show that joint characterization is capable of detecting and isolating signals which are not evident from either PCA or t-sne alone. Broadly, t-sne is effective at rendering a randomly oriented low-dimensional map of local clusters, and PCA renders this map interpretable by providing global, physically meaningful structure. This approach is illustrated using imaging spectroscopy data, and may prove particularly useful for other geospatial data given robust local variance structure due to spatial autocorrelation and physical interpretability of global variance structure due to spectral properties of Earth surface materials. However, the fundamental premise could easily be extended to other high dimensional datasets, including image time series and non-image data.

Introduction

The challenge of high dimensional data analysis has been recognized for decades as critical for scientific progress (Donoho, 2000). Physical, chemical, biological, geological, and astronomical instruments routinely collect measurement streams which rapidly reach terabytes in volume. The Large Synoptic Survey Telescope (under construction at time of writing) is expected to collect 20 terabytes of raw data each night using a 3.2 gigapixel camera (Kahn et al., 2010); a single future NASA mission for the Surface Biology and Geology designated observable is planned to collect – and downlink to Earth – roughly 100 terabytes per day (Jenkins et al., 2019). Similar missions are planned by others, including the European Space Agency and the private sector. As such data continue to grow in size and complexity, novel algorithms are needed to extract relevant signals and mitigate the effects of noise.

In high dimensional data, information can exist at different scales. Sometimes, information is present in *global patterns* – the overall structure of the data based on all the observations at once. Other times, additional information can be found in *local structure* – often conceptualized as connectivity or clustering relations among similar observations. In such cases, the meaningful signal may be well-represented as a low-dimensional manifold embedded in a much higher dimensional data space. Given the wide range of scales at which meaningful information may exist, it is plausible that the same dataset may possess useful information at both global and local scales. This situation, which we refer to as multiscale structure, may even be quite common.

Different analysis tools are designed to capture different scales of variability. The well-established tool of Principal Component Analysis (PCA; (Pearson, 1901)) can be considered to reflect the global variance structure of a dataset because it uses the correlation structure of all observations at once to apply a single, linear transform to the entire dataset. In contrast, more recent manifold learning algorithms have sought to capture or retain local structure, for instance by using nearest neighbor-type approaches (Cayton, 2005). Both approaches have their merits. Both can be useful analytic tools in some circumstances, and not useful in other circumstances, depending on the variance structure of the data.

We propose a new way to consider high dimensional data structures – through *joint characterization*. Rather than only using one or another type of analysis, joint characterization seeks to exploit synergies between multiple scales of variance and proximity to capture multiscale signals that are not well-represented by either tool on its own. These signals can be manifest as structure in the space of the bivariate (global vs local) distributions.

While the fundamental idea can be generalized, for the purposes of this study we illustrate the approach using a progression of low to high dimensional image data. To do so, we choose two commonly used dimensionality reduction tools: PCA as our global projection; and t-stochastic neighbor embedding (t-sne; (van der Maaten & Hinton, 2008)) as our local algorithm. We explore a series of increasingly complex examples to illustrate the strengths of this approach, starting with highly simplified synthetic images and progressing to real-world airborne imaging spectrometer data. Imaging spectroscopy data are a highly relevant potential application for the approach due to their high dimensionality and the status of imaging spectroscopy as an emerging area of remote sensing research (as indicated by the planned missions referenced above).

Joint characterization is shown to consistently harness the strengths of both methods, using the interpretability of low-order PCs to orient the well-separated but disordered clusters that are the hallmark of t-sne. The utility of joint characterization becomes increasingly apparent as data become more complex and multiscale structure becomes more important. In the imaging spectroscopy dataset, joint characterization is shown to be capable of identifying subtle differences in reflectance spectra of green vegetation at biophysically meaningful wavelengths. The structure captured by joint characterization is not readily apparent in either the PC or t-sne spaces alone, indicating clear potential utility of the approach for imaging spectroscopy applications. Joint characterization may prove particularly useful for geospatial datasets. In such data, spatial autocorrelation can translate into physically meaningful variance at local scales, and the finite range of reflectance of Earth surface properties can translate into consistent, physically interpretable patterns of global variance.

Analysis

1. A Simple Synthetic Image

Simplified examples can help illustrate the approach. We begin with a synthetic image comprised only of shades of gray (Figure 1). This 100 x 100 pixel 8-bit image is divided into a 10 x 10 grid of monoshade chips, varying linearly from black (data values [0, 0, 0]) to white (data values [255, 255, 255]). Every pixel within each chip possesses identical data values. Because each pixel's data values are identical across the three channels, each 3-D [r,g,b] vector can be represented by a single scalar value with no loss of information.

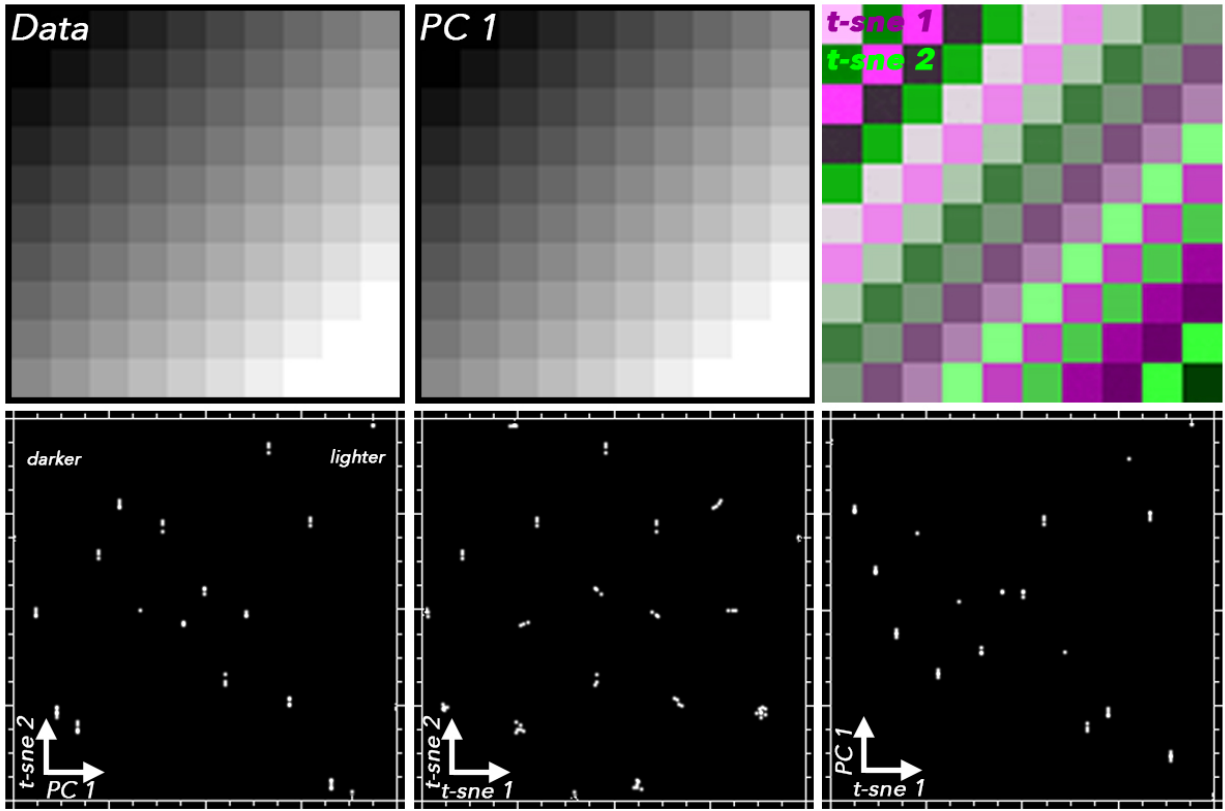

*Figure 1. A synthetic grayscale image illustrates how PCA can orient t-sne clusters. In the case of a grayscale image (top left), continuous variation of one principal component (top center) captures all the information in the data. 2D t-sne (top right) correctly assigns each equal-brightness matrix diagonal to a small range of t-sne values. The t-sne space (bottom center) has well-separated clusters corresponding to discrete brightness values, but conveys no information about the global spatial structure (brightness gradient) of the matrix. The joint PC + t-sne space (lower left and lower right) retains the local cluster structure, but orients the global relationships among clusters to align them along the diagonal brightness gradient.*

The geometric implication of all 3 channels of the image possessing equal values for all pixels is that the data lie on a 1-D line within a redundant 3-D feature space. Standard PCA reflects this geometric arrangement by partitioning 100% of the total variance (within numerical noise) into the first PC. As a result, only one dimension of PCA is needed to exactly replicate the original image data.

The t-sne output in this simplified example (bottom center) illustrates several aspects of the method. First, the algorithm separates the data into approximately equally spaced clusters. Each cluster of points in the t-sne scatterplot corresponds to a homogenous diagonal of the matrix of color chips – i.e., all points in the image with a single [r,g,b] value.

This leads to the second important point illustrated in Figure 1 - that the relative positions of the clusters in t-sne space are arbitrary. This property is a significant departure from PCA, where position along each axis corresponds to relative contribution of a specific eigenvector. Not so for t-sne.

The third and fourth salient properties are less obvious at first glance: clusters of identical values have nonzero dispersion in t-sne space, and different t-sne runs with identical hyperparameters give slightly different results. These are also in contrast to PCA, where data values correspond to one and only one set of PC values.

We show results using the scikit-learn default hyperparameter settings here (most importantly, perplexity = 30), but note that the experimentation suggests the general structure of the output is robust to a wide range of hyperparameter choices.

Examining the PC + t-sne outputs together – joint characterization – retains important information from each. PCA effectively orients the data along its global variance structure – in this case, from lowest to highest brightness values. T-sne effectively partitions similar (in this case, identical) values into well-separated clusters, but does not structure these clusters in a meaningful way. By combining the global and local information captured by PCA and t-sne, both sets of information are retained and we are able to illustrate patterns which would be hidden by one or the other tool on its own. In the 1-D case of Figure 1 there is no local structure and the $1^{st}$ PC captures all the global structure, so joint characterization provides no new information. This is not the case for the datasets with greater dimensionality examined below.

2. A More Complex Synthetic Image: Discrete 3D

Further insight can be obtained by examining a slightly more complex synthetic dataset (Figure 2). Here, three additional submatrices are introduced which illustrate binary mixing for each of the (red, green); (red, blue); and (green, blue) combinations of primary additive colors.

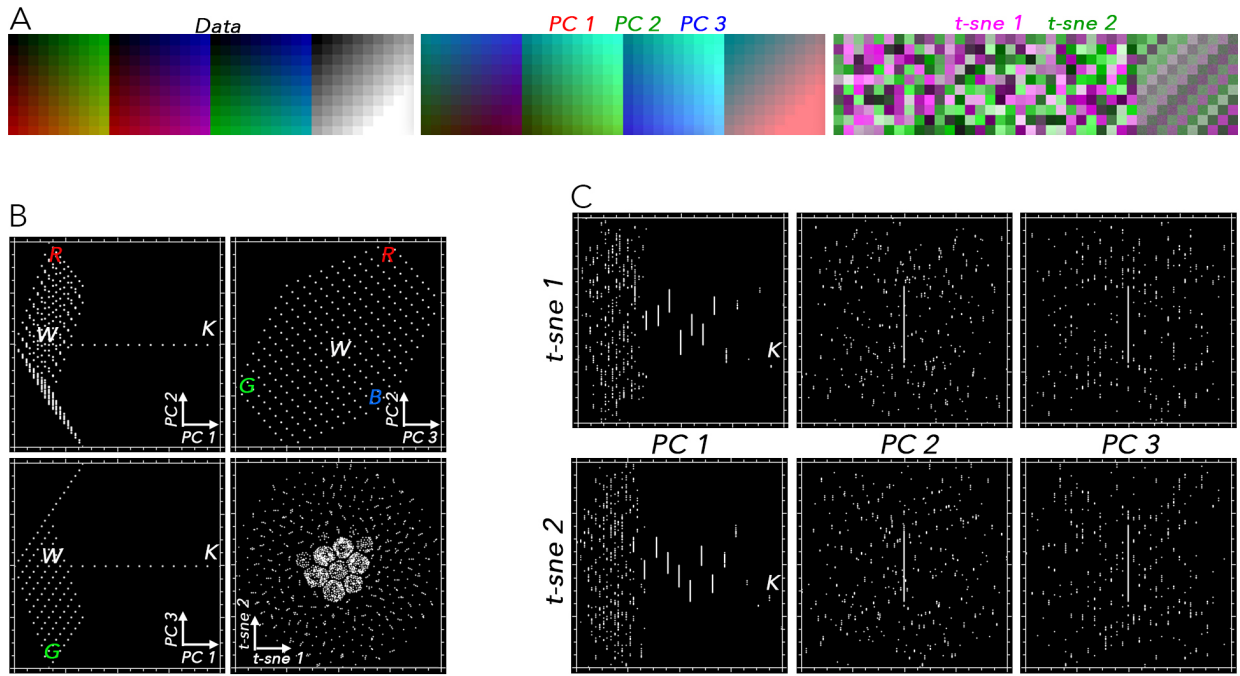

*Figure 2. Synthetic binary color mixtures illustrate local and global information. A synthetic image (A; left) comprised of submatrices with both binary color mixtures and grayscale brightness variability is analyzed with both PCA (A; center) and t-sne (A; right). Three PCs are now required to describe the RGB data. In PC space (B), the binary color mixtures form a hexagonal manifold (PC 2 vs 3) arranged by hue and a perpendicular gray axis to represent brightness. In t-sne space (B; lower right), each equal-valued diagonal of the grayscale submatrix forms a large cluster in the center while each distinct color of the binary mixing submatrices forms a denser cluster in the surrounding cloud. The joint PC+t-sne space (C) aligns the randomly oriented t-sne clusters along directions of global variance (brightness), with the PC 1 projection clearly separating the grayscale and color submatrices, and PC 2 and 3 projections orienting clusters in the color submatrix on the basis of hue.*

PCA in this case yields markedly more structure to the data. PC 1 corresponds to overall pixel brightness and PCs 2 and 3 form a discrete, hexagonal approximation of the color wheel, where each of the 6 vertices corresponds to either a (red, green, blue) primary color or a (yellow, cyan, magenta) binary mixture. The gray axis protruding from the center of the color wheel (white (W)) towards high PC 1 values (black (K)) corresponds to the grayscale submatrix we have seen in Figure 1, which again can be completely represented by a single dimension.

The t-sne result for this more complex synthetic image bears notable similarities to the t-sne result for the grayscale-only image. Again, each set of pixels with identical values forms a cluster. The increased number of unique [R,G,B] values is reflected in the increased number of clusters in t-sne space. Again, the clusters form an approximately equidistant lattice-like structure; again, the relative position of clusters within this structure is arbitrary; again, pixels with exactly identical [R,G,B] values (gray) form clusters with nonzero variance. The grayscale submatrix occupies the central portion of the t-sne plot and differs from the binary mixing submatrices because it has a greater number of pixels with identical data values (entire submatrix diagonals, rather than individual matrix elements).

As the complexity and dimensionality increase from Figure 1 to Figure 2, the value added by joint characterization becomes apparent. The global variance structure captured by PCA serves to orient the local structure captured by t-sne. However, in this image, the homogeneity of each color chip in this image limits the usefulness of local structural metrics like t-sne. As we shall see below, local information can be much more important as image complexity increases further.

3. Greater Complexity: Continuous 3D with Compression Artifacts

While simple datasets comprised of clusters with precisely equal data values can provide valuable insight, most real-world images contain continuously varying data values. Figure 3 illustrates PCA, t-sne, and the joint characterization approach for a third example image with both discrete and continuous variability. Importantly, the image has also undergone lossy compression using the commonly used JPEG algorithm. Lossy compression is typically avoided for datasets with scientific applications because it changes underlying data values. However, it does so in a way designed to be difficult for the human eye-brain system to perceive. For this reason, ability to detect these compression artifacts can be an indicator of the power of an analysis approach to extract subtle, low-variance information which might be difficult for humans to visualize in the raw data.

Again, as image complexity increases further, the relevance of the joint characterization approach becomes clearer. PCA again reflects the global variance structure of the data – in this case, a 3-D color cube. Locally, t-sne again partitions the sets of identical or nearly identical values of the color chips in the bottom half of the image to tight clusters in the periphery of the plot. The continuum in the top half of the data image is expressed as a more tortuous central point cloud. T-sne appears to be sufficiently sensitive to coherent small amplitude variability to capture JPEG compression artifacts which are not obvious to the naked eye. Together, the joint space retains benefits of both PC and t-sne: global ordering and local cluster separability, respectively.

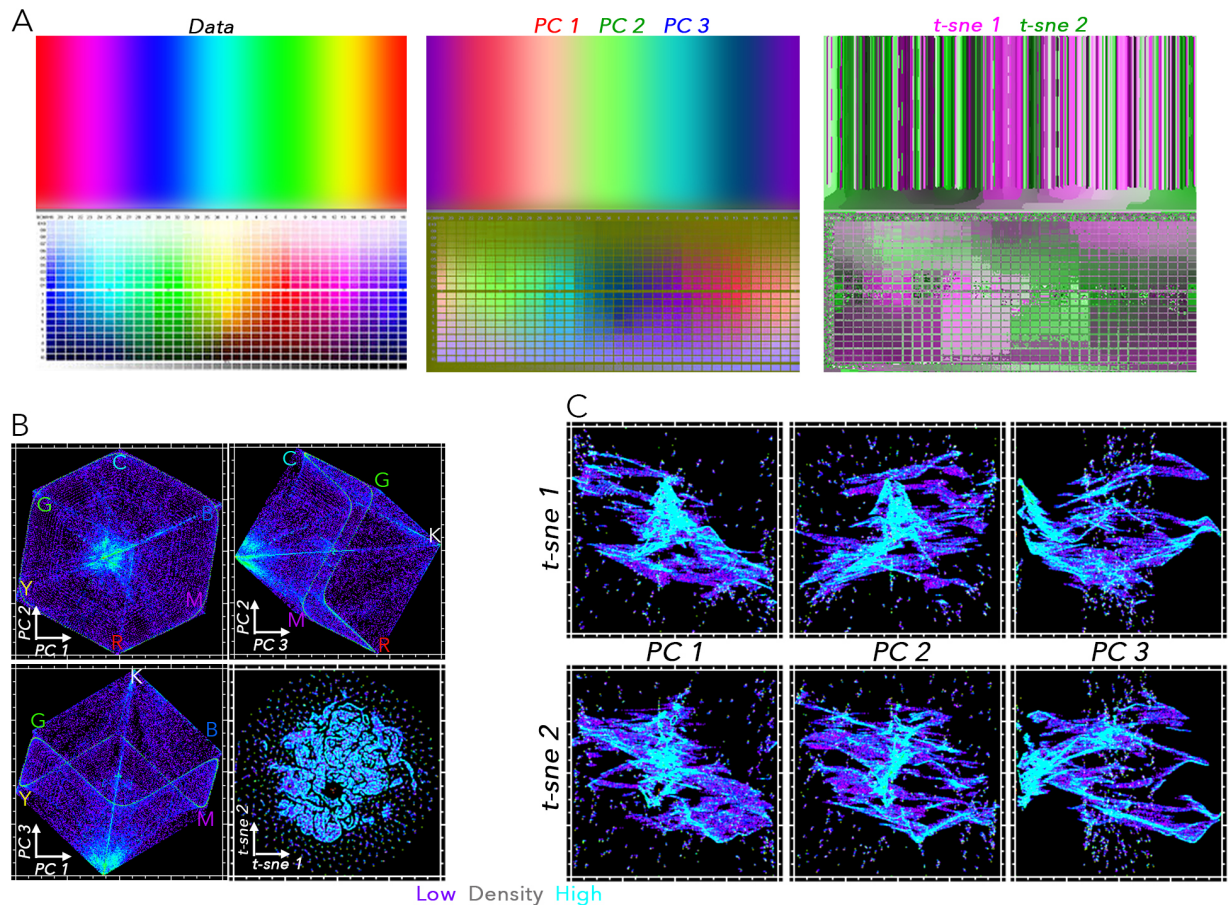

*Figure 3. Joint characterization reveals continuous and discrete structure in a more complex 3D example. Both continuous and discrete variability is present in these discrete and continuous [R,G,B] color tables (A; left). PCA (A; center) retains the global color gradients while t-sne (A; right) separates colors discontinuously on the basis of local structure. In PC space (B) distinct [R,G,B] values fill out a color cube with a gray axis spanning the diagonal between black and white. The t-sne space (B) finds subtle nonlinear manifolds that appear to correspond to compression artifacts in the continuous hue scale, surrounded by tight clusters that correspond to specific [R,G,B] triplets on the discrete array. The joint PC+t-sne characterization (C) introduces complexity by disrupting the interpretable geometric structure of the PC color space with the apparently random structure of the t-sne space.*

4. Greater Dimensionality and Complexity: Airborne Imaging Spectroscopy

One promising application for joint characterization is in the examination of Earth-observing images from sensors on aerial and satellite platforms. Imaging spectroscopy is a particularly interesting use case because it allows us to explore images with much greater dimensionality than the images recorded by typical cameras. Whereas a digital camera images the visible spectrum using 3 color channels (red, green and blue primary colors), imaging spectrometers image a much wider range of visible to Shortwave Infrared (SWIR) wavelengths with hundreds of narrowband spectral channels. The result is a hyperspectral image cube where each geographic pixel provides a very detailed optical reflectance spectrum capable of resolving subtle molecular absorptions characteristic of different chemical compositions of the material being imaged. The broad scale global structure of each spectrum is characteristic of the material type (e.g. soil, vegetation, rock, water, snow) while the finer scale absorption features allow differentiation within each material type (e.g. soil composition, vegetation type and condition, rock mineral content). The resulting spectral feature space generally has >100 dimensions corresponding to each of the narrow spectral channels in the image cube. PCA is commonly used to characterize the lower-D global structure of the hyperspectral feature space with respect to different types of materials, but fine scale absorption features exist which contribute much less variance than the overall spectral shape and so tend to form amorphous continua embedded within the low-D subspace of the low order PCs. The situation is analogous to the continuous color scale in Figure 3 in which the primary colors control the overall cubic form of the 3D PC space, while the continuum of additive colors form continua on the manifold between the primary color apexes. The main difference is the vastly greater information content of imaging spectroscopy, resulting from narrower spectral channels (e.g. ~40 channels in the visible spectrum alone) and much broader spectral range (e.g. 400 nm (visible blue) to 700 nm (visible red) to 900 nm (Near Infrared) to 2500 nm (Shortwave Infrared)).

In the context of this expository analysis, industrial-scale agriculture is a particularly well-posed problem because of the easily comprehensible geographic patterns that occur at spatial scales coarser than pixel resolution. Whereas individual fields are generally planted with a single crop type having a similar overall spectral shape, spatial variations in soil moisture and nutrients within individual fields can result in subtle differences in the reflectance of the aggregate spectra corresponding to the area imaged by each pixel. The ability to image these subtle spectral features is part of the basis for the great potential of imaging spectroscopy to precision agriculture and crop monitoring. To date, over 23 TB of such imagery has been collected by NASA's Airborne Visible / Infrared Imaging Spectrometer (AVIRIS) and are publicly available free-of-charge at: https://aviris.jpl.nasa.gov/dataportal/.

We use an example AVIRIS hyperspectral cube of agriculture in the California's Central Valley, home to extensive industrial agriculture with total cash receipts exceeding $50 billion/year (CDFA, 2019). The area shown in Figure 4 is a 500 x 500 pixel (10 x 10 km) subset of flightline f140606t02p00r04. Each pixel represents a spatially integrated optical reflectance spectrum at roughly 16.5 m spatial resolution and 10 nm spectral resolution over the range 365 to 2496 nm. Bands 1-14, 49-64, 75-84, 90-120, 150-180, and 210-224 were excluded due to atmospheric absorptions and/or sensor noise. 107 bands were retained for analysis.

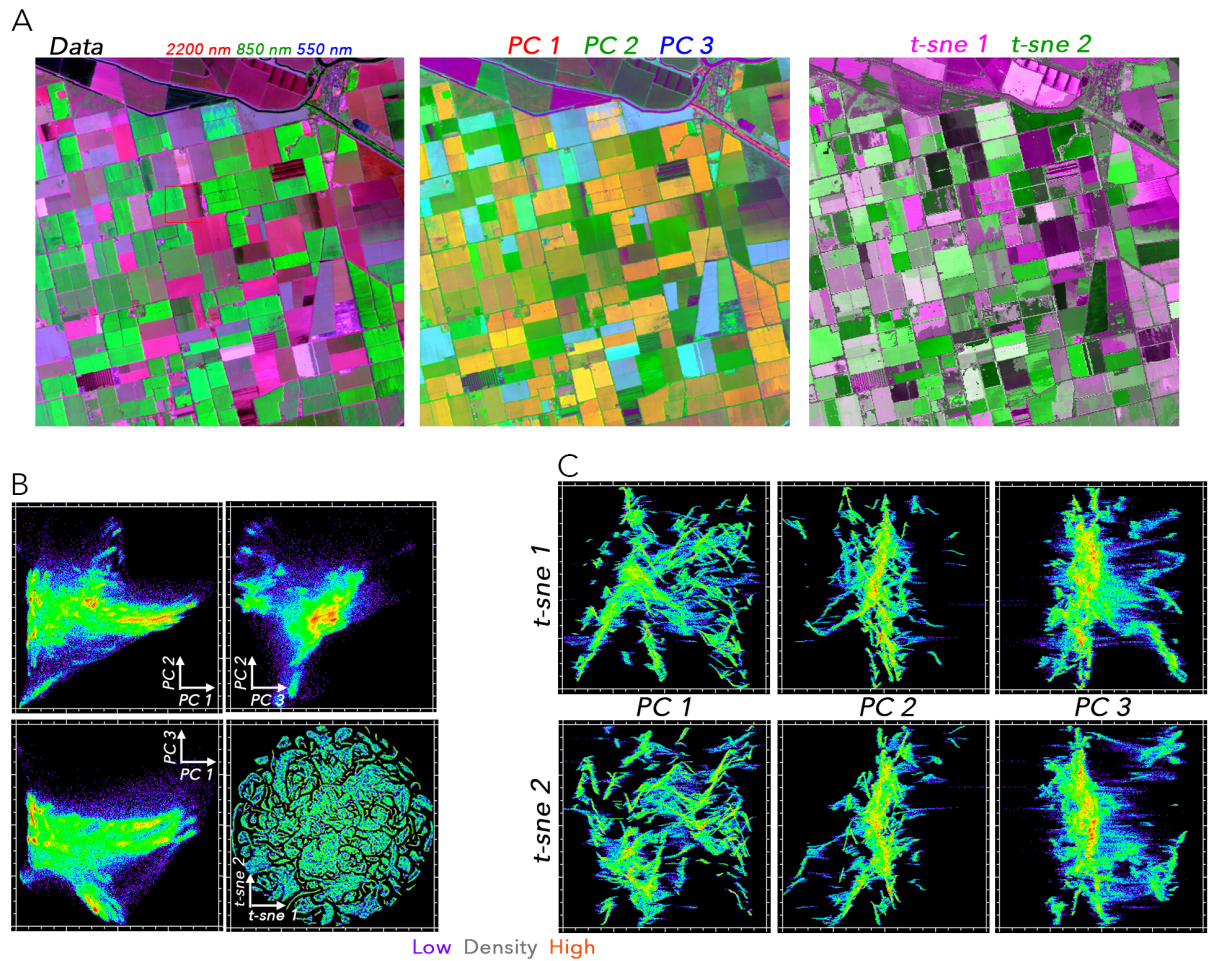

*Figure 4. Joint characterization reveals novel structure in imaging spectroscopy data. The data here (A; left) show an agricultural landscape in California's Central Valley as imaged by the 224-channel AVIRIS hyperspectral sensor. PCA (A; center) represents the global variance continuum, and t-sne (A; right) captures local variance. The structure of the data in PC space (B) distinguishes differences in vegetation, soil and water reflectance while the structure in t-sne space (B; lower right) clusters the data but does not orient according to any global structure. Joint characterization (C) contains both the global variance structure represented by PCA and the local structure found by t-sne, effectively segregating the continuous PC space into many distinct clusters and gradients.*

The AVIRIS dataset clearly contains real-world complexity. However, the same properties which characterize the synthetic datasets are also evident even here. The 3 low-order PCs effectively capture global variance structure, capturing >97% of variance in this dataset and clearly grading between distinct materials like soils, senescent vegetation, green vegetation, and standing water. This single subset has similar global variance features to those found in previous, more extensive observations of agricultural landscapes across the region in (Sousa & Small, 2018).

Even in the face of real-world complexity, t-sne can be remarkably effective at capturing local structures in imaging spectroscopy data. Clusters in t-sne space unambiguously correspond to spatially coherent patterns of intra- and inter-field variability. However, as in the synthetic images, the t-sne space is clustered but unstructured. Adjacent clusters generally have no clear relation to each other, and clusters in disparate corners of the t-sne space can possess great similarity. This property is inherent to t-sne and limits its utility for imaging spectroscopy data – unless additional information is introduced.

Joint characterization provides this additional information. In the case of these data, the joint space clearly both retains useful clustering from t-sne, and also orients those clusters using the global variance structure captured by PCA. As a result, subtle but consistent interpretable differences can be isolated within the imaging spectroscopy data which could not readily be identified from global or local characterization alone.

Figure 5 shows an example using 10 regions of interest (ROIs) from the PC+t-sne space. These ROIs are chosen to illustrate the observed local variability in reflectance of green vegetation spectra. All green vegetation spectra in the image cluster in one corner of the PC space (lower left scatterplot), and form distinct clusters throughout the t-sne space (inset lower left), but form clearly separated clusters across the range of t-sne values in the joint characterization space (upper left scatterplot). Mean reflectance spectra of each ROI are plotted in the lower right. All have a similar overall shape (associated with their proximity in PC space) with characteristic absorptions typical of green vegetation, but differ interpretably at wavelengths corresponding to known biophysical absorption features like leaf water, cellulose/lignin, and photosynthetic pigments. Differences among ROIs are easier to visualize when accentuated by differencing pairs of mean spectra (inset right). The differences among the mean spectra are also indicated by the varying amplitude sequences in each wavelength interval (inset labels). The variations in SWIR (1500 nm to 2500 nm) amplitude are primarily a result of differences in depth of the liquid water absorptions in the $H_2O$ gaps driven by differences in leaf water. It is notable that the spectral subsets of every pair of ROIs has Transformed Divergence of 2.0, indicating complete spectral separability in the original 224-dimensional feature space of the hyperspectral cube. When the pixels in each contiguous ROI from the PC+t-sne space are plotted in their corresponding locations in geographic space (upper right image) it is immediately apparent that each ROI cluster from PC+t-sne space corresponds to a distinct, spatially contiguous region within individual or adjacent fields in geographic space. This spatial contiguity is consistent with the local structure revealed by the PC+t-sne space being related to crop-specific features in the reflectance spectra. These subtle intra-field variations in reflectance can also be seen on the sides of the hyperspectral cube in Figure 5.

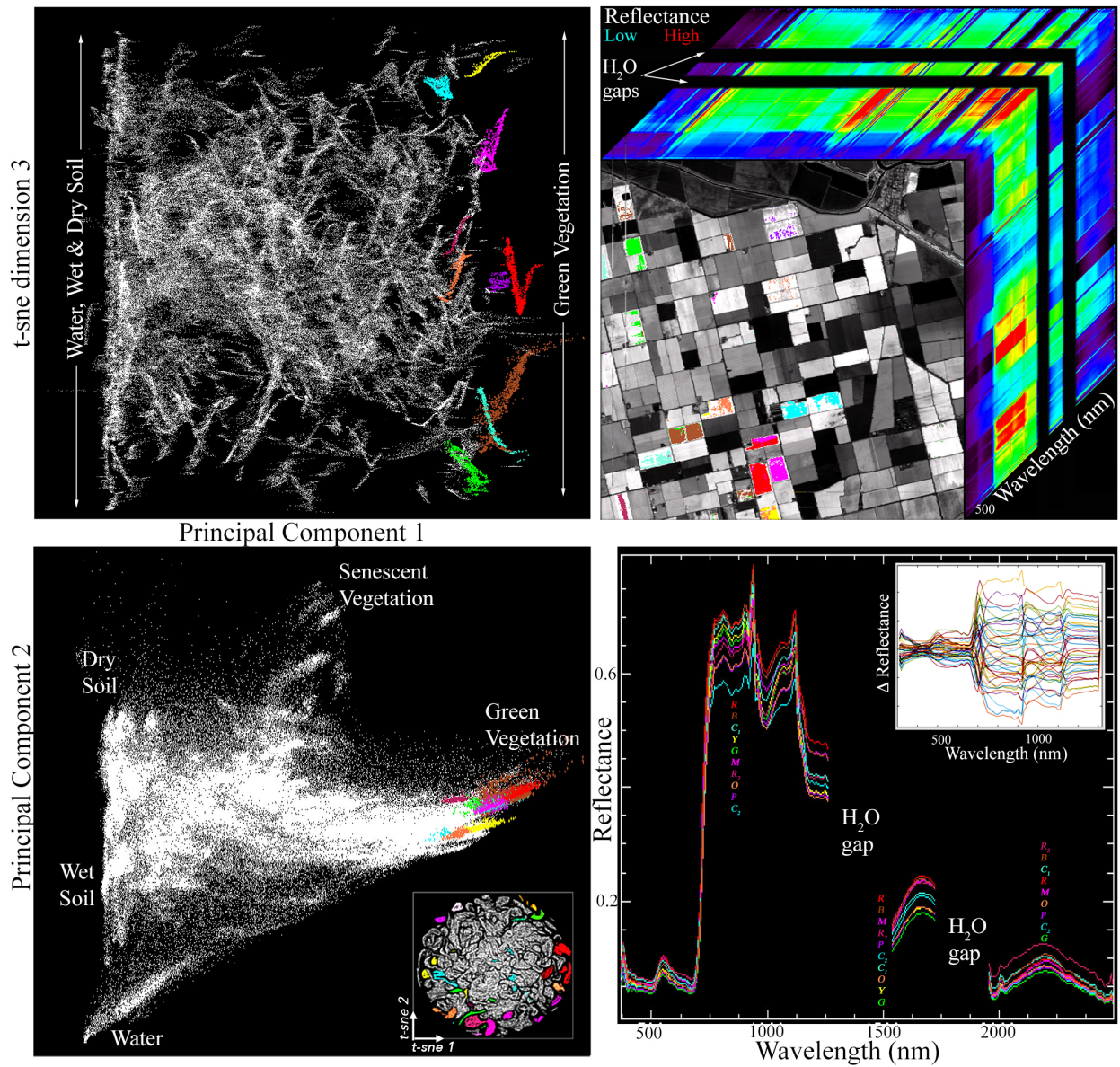

*Figure 5. Multiscale spectral feature space of an AVIRIS hyperspectral image cube of agricultural plots in the Great Central Valley of California. The spectral feature space of PCs 1 and 2 (lower left) shows global structure related to characteristic reflectance spectra of distinct land cover types. The complementary feature space of PC1 and t-sne1 (upper left) shows finer scale clustering related to local structure corresponding to subtle differences among similar land cover spectra. Distinct clusters (colored) of spectra generally correspond to spatially contiguous patches within individual plots (upper right) while the mean spectra of each cluster (lower right) all correspond to green vegetation, but with varying differences in amplitude and curvature related to varying depth of H2O absorption bands resulting from varying leaf water content of different crops. Pairwise differences of cluster mean reflectances (inset) illustrate wide range of varying spectral slopes indicative of differing H2O absorption strength, despite similar shapes of vegetation spectra that result in cluster overlap in PC1/PC2 feature space.*

Discussion

From Figure 4 and Figure 5, it is apparent that the joint characterization approach can be useful for analysis of imaging spectroscopy datasets, at least of agricultural landscapes. While some of the features of imaging spectroscopy data may render it particularly amenable to this type of analysis, the approach is highly general and might easily be extended to other high dimensional datasets. Potential implications and avenues for future work are discussed below.

1. Importance of both local & global variance

The foundation of this study is the recognition that the information content of high dimensional datasets can map onto multiple scales of variance. In essence, some data are too rich to be well-described by variation measured at a single scale. In the case of imaging spectroscopy, global and local scales are frequently recognized as the "spectral continuum" and "absorption features" (respectively), and so have established physical attributes (Clark & Roush, 1984). The global variance structure of Earth surface reflectance also has a well-understood physical meaning in terms of a mixing space of spectrally distinct endmember materials (Kauth & Thomas, 1976; Crist & Cicone, 1984; Small, 2004; Small & Milesi, 2013; Sousa & Small, 2017, 2018). This may not generally be the case for all types of data.

2. Implications for the remote sensing community

Imaging spectroscopy is an emerging subdiscipline in remote sensing. After decades of instrument development, and early successes like Hyperion (Pearlman et al., 2003) and HICO (Corson et al., 2008), a new generation of satellite missions are at or nearing operational status. While multispectral satellite imagery has been collected for over 40 years, image analysis methods must continue to evolve to match the higher dimensional demands of hyperspectral data. Tools like joint characterization will be necessary for the scientific community to fully exploit the information captured by these data.

An additional promising area of inquiry is in the realm of spatiotemporal analysis. Temporal feature spaces are inherently higher dimensional than spectral feature spaces because they are a result of temporal progressions of physical states less constrained by the finite number of material properties and molecular structures that give rise to featue spaces of reflectance spectra (Small, 2012). It is likely that image time series may also be characterized by signals existing at multiple scales, especially given the wide range of ways that Earth surface reflectance can evolve through time (vegetation phenology, fire, flooding, land cover change, etc.).

3. Open questions for future work

Several avenues for future work exist. First, we use PCA and t-sne in this analysis as tools for measuring global and local variance, but others exist. Which, if any are most effective? Using what criteria, and under what circumstances? Second, t-sne (and other manifold learning approaches) can yield variable results. How sensitive is the result to hyperparameter choice? How well do clusters observed by joint characterization align with physical properties of the data? In what, if any, cases are clusters spurious, or have no clear physical meaning? Questions like these, and many others, will be critical to examine in detail as joint characterization is applied to a greater diversity of data.

Conclusions

We introduce the idea of joint multiscale characterization. The underlying principle is that information in high dimensional data can exist at multiple scales of variance. Joint characterization uses global and local variance metrics together to identify data patterns which are not evident from either metric alone. We illustrate this principle using Principal Components Analysis as a global variance metric and t-stochastic neighbor embedding as a local variance metric. Examples begin with low dimensional synthetic image data and progress in complexity and dimensionality to airborne imaging spectroscopy data of an agricultural landscape. The utility of joint characterization increases as data complexity increases, culminating in clear separability of green vegetation spectra which differ at wavelengths corresponding to key biophysical traits. Joint characterization may prove particularly useful for geospatial datasets given both robust local structure (due to spatial autocorrelation) and physical meaning of global variance (due to spectral mixing), but the idea could be generalized to other data types as well.

Acknowledgement

D. Sousa's work was done as a private venture and not in the author's capacity as an employee of the Jet Propulsion Laboratory, California Institute of Technology.  C. Small acknowledges the support of the Lamont Doherty Earth Observatory of Columbia University. 

Supplementary Information

1. T-sne Hyperparameter Sensitivity

Figure S1 illustrates the effect of varying perplexity on the trial dataset from Figure 2. Here, perplexity is varied across the range of suggested values (from 5 to 50, in increments of 5). The overall effect of changing perplexity in this dataset is on the dispersion of the central grayscale clusters. In general, t-sne clusters comprised of equal-valued grayscale pixels are most compact and most separated from each other at low perplexity values. As perplexity increases, the dispersion of the clusters increases and they grow towards each other, ultimately filling the majority of the space of the central portion of the plot. Interestingly, this occurs only for the clusters corresponding to grayscale pixels, and not clusters corresponding to binary color mixtures. Other key hyperparameters (early exaggeration, learning rate, number of iterations) were varied across their suggested ranges as well, but none demonstrated any appreciable sensitivity comparable to that of perplexity.

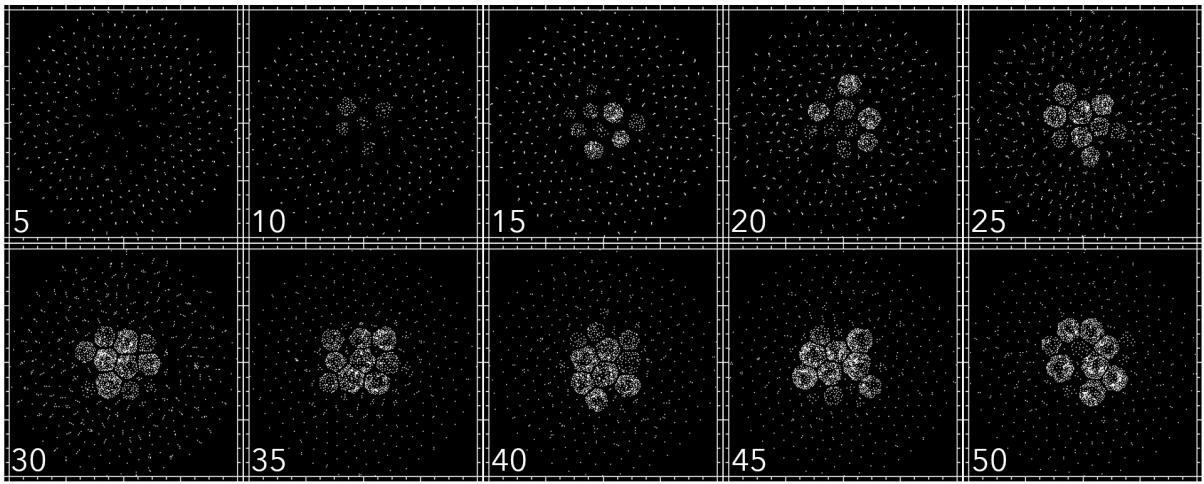

*Figure S1. The t-sne hyperparameter of perplexity impacts cluster dispersion. As the t-sne hyperparameter of perplexity increases across its recommended range of values, the central clusters corresponding to equal-valued grayscale matrix diagonals expand to fill the center of the plot. The outer clusters corresponding to binary color mixtures also expand but show considerably less sensitivity.*